\documentclass[lettersize,journal]{IEEEtran}
\usepackage{algorithm}
\usepackage{array}
\usepackage[caption=false,font=normalsize,labelfont=sf,textfont=sf]{subfig}
\usepackage{textcomp}
\usepackage{stfloats}
\usepackage{url}
\usepackage{verbatim}
\usepackage{graphicx}
\usepackage{cite}
\usepackage{amsmath,amssymb,amsfonts}
\usepackage{algorithmic}
\usepackage{times}
\usepackage{latexsym}
\usepackage{makecell}

\usepackage{here}
\usepackage{stmaryrd}
\usepackage{multirow}
\usepackage{changepage}
\usepackage{threeparttable}
\usepackage{tabu}
\usepackage{longtable}
\usepackage{float}
\usepackage{colortbl}
\usepackage[dvipsnames,table]{xcolor}
\usepackage{tabularx}

\usepackage{array}
\usepackage[utf8]{inputenc} 
\usepackage[T1]{fontenc}    
\usepackage[hidelinks]{hyperref}       
\usepackage{booktabs}       
\usepackage{nicefrac}       
\usepackage{microtype}      
\usepackage{lipsum}
\begin{document}
\bstctlcite{BSTcontrol}

\title{Calibration of Transformer-based Models for Identifying Stress and Depression in Social Media}

\author{Loukas Ilias, Spiros Mouzakitis, Dimitris Askounis
\thanks{Manuscript received 5 January 2023; revised 28 March 2023 and 6 May 2023; accepted 2 June 2023. \textit{(Corresponding author: Loukas Ilias.)}}
\thanks{The authors are with the Decision Support Systems Laboratory, School of Electrical and Computer Engineering, National Technical University of Athens, 15780 Athens, Greece (e-mail: lilias@epu.ntua.gr; smouzakitis@epu.ntua.gr; askous@epu.ntua.gr).}}

\markboth{Journal of \LaTeX\ Class Files,~Vol.~14, No.~8, August~2021}%
{Shell \MakeLowercase{\textit{et al.}}: A Sample Article Using IEEEtran.cls for IEEE Journals}


\maketitle

\begin{abstract}
In today's fast-paced world, the rates of stress and depression present a surge. 
People use social media for expressing their thoughts and feelings through posts. 
Therefore, social media provide assistance for the early detection of mental health conditions. 
Existing methods mainly introduce feature extraction approaches and train shallow machine learning classifiers. 
For addressing the need of creating a large feature set and obtaining better performance, other researches use deep neural networks or language models based on transformers.
Despite the fact that transformer-based models achieve noticeable improvements, they cannot often capture rich factual knowledge. Although there have been proposed a number of studies aiming to enhance the pretrained transformer-based models with extra information or additional modalities, no prior work has exploited these modifications for detecting stress and depression through social media.
In addition, although the reliability of a machine learning model’s confidence in its predictions is critical for high-risk applications, there is no prior work taken into consideration the model calibration. 
To resolve the above issues, we present the first study in the task of depression and stress detection in social media, which injects  extra linguistic information in transformer-based models, namely bidirectional
encoder representations from transformers (BERT) and MentalBERT.  Specifically, the proposed approach employs a Multimodal Adaptation Gate for creating the combined embeddings, which are given as input to a BERT (or MentalBERT) model.
For taking into account the model calibration, we apply label smoothing. 
We test our proposed approaches in three publicly available datasets and demonstrate that the integration of linguistic features into transformer-based models presents a surge in the performance. 
Also, the usage of label smoothing contributes to both the improvement of the model's performance and the calibration of the model. 
We finally perform a linguistic analysis of the posts and show differences in language between stressful and non-stressful texts, as well as depressive and non-depressive posts.
\end{abstract}


\section{Introduction}
According to the World Health Organization (WHO) \cite{stress_definition}, stress can be defined as any type of change that causes physical, emotional or psychological strain. Stress comes with a number of categories \cite{types_of_stress}, namely physical, psychological, psychosocial, and  psychospiritual stress. Excessive stress can lead to anxiety disorders or even depression. Depression entails a great number of symptoms, including loss of interest, anger, pessimism, changes in weight, feelings of worthlessness, thoughts of suicide, and many more. According to the WHO \cite{depression_definition}, around 280 million people in the world have depression. China, India, the United States, Russia, Indonesia, and Nigeria are some of the countries presenting the highest rates of depression \cite{depression_rates}. People with stress and depression use social media platforms, including Twitter and Reddit, and share their thoughts, emotions, feelings, and so on through posts or comments with other users. Therefore, social media constitute a valuable form of information, where linguistic patterns of depressive/stressful posts can be investigated.

Existing research initiatives exploit social media data for identifying depressive and stressful posts. The majority of these research works \cite{8681445,Chandra_Guntuku_Buffone_Jaidka_Eichstaedt_Ungar_2019} employ feature extraction approaches and train shallow Machine Learning (ML) algorithms.  Employing feature extraction approaches constitutes a tedious procedure and demands domain expertise, since the authors may not find the optimal feature set for the specific problem. At the same time, the train of shallow ML algorithms does not yield optimal performance and does not generalize well to new data. For addressing these limitations, other approaches \cite{FIGUEREDO2022100225} use deep neural networks, including Convolutional Neural Networks (CNNs),  bidirectional long short-term memory (BiLSTM), and so on, or transformer-based networks. In addition, there are researches employing ensemble strategies \cite{9733425}. However, these approaches increase substantially the training time, since multiple models must be trained separately. In addition, recently there have been studies \cite{poerner-etal-2020-e,kassner-schutze-2020-negated} showing that transformer-based models struggle or fail to capture rich knowledge. For this reason, there have been proposed methods for enhancing these models with external information or additional modalities \cite{peinelt-etal-2021-gibert-enhancing,wang-etal-2021-k,10.5555/3454287.3454289,peters-etal-2019-knowledge}. However, existing research initiatives in the tasks of stress and depression detection through social media have not exploited any of these approaches yet. In addition, the reliability of a machine learning model’s confidence in its predictions, denoted as calibration \cite{doi:10.1080/01621459.1982.10477856,VerificationofProbabilisticPredictionsABriefReview}, is critical for high risk applications, such as deciding whether to trust a medical diagnosis prediction \cite{doi:10.1177/0962280213497434,10.1136/amiajnl-2011-000291,pmlr-v97-raghu19a}.  Although methods regarding the confidence of models' predictions have been introduced in many studies, including suicide risk assessment \cite{sawhney-etal-2022-risk}, sleep stage classification \cite{9570807}, and so on, no prior work for stress and depression detection has taken into account the level of confidence of models' predictions, creating in this way overconfident models.

To tackle the aforementioned limitations, we propose a method, which injects  extra linguistic information into transformer-based models, namely BERT and MentalBERT. Firstly, we extract various linguistic features, including NRC Sentiment Lexicon, features derived by Latent Dirichlet Allocation (LDA) topics, Top2vec, and Linguistic Inquiry and Word Count (LIWC) features.  Regarding the LDA topic-based features, this is the first study in terms of the tasks of stress and depression detection via social media texts utilizing the Global Outlier Standard Score (GOSS) \cite{liu-etal-2016-detecting}, which captures the text's interest on a specific topic in comparison with other texts. After passing each text through a transformer-based model, we project the linguistic information to the same dimensionality with the outputs of the transformer models. Next, we concatenate the representations obtained by BERT (or MentalBERT) and linguistic information and apply a Multimodal Adaptation Gate \cite{rahman-etal-2020-integrating}, where an attention gating mechanism is used for controlling the importance of each representation. Similarly to \cite{jin-aletras-2020-complaint}, we modify M-BERT \cite{rahman-etal-2020-integrating} by replacing the multimodal information with linguistic information. Finally, a shifting component is exploited for calculating the new combined embeddings. The new combined embeddings are passed through a BERT (or MentalBERT) model, where the classification [CLS] token is fed to Dense layers for getting the final prediction. In addition, for preventing models becoming too overconfident, we use label smoothing. According to  Müller et al. \cite{NEURIPS2019_f1748d6b}, label smoothing has been used successfully to improve the accuracy of deep learning models across a range of tasks, while at the same time it implicitly calibrates learned models so that the confidences of their predictions are more aligned with the accuracies of their predictions. We use metrics for assessing both the performance and the calibration of our model. We also demonstrate the efficiency of label smoothing in both calibrating and enhancing the performance of our model. We test our proposed approaches on three publicly available datasets, which differentiate \textit{(i)} stressful from non-stressful texts, \textit{(ii)} depressive from non-depressive posts, and \textit{(iii)} posts indicating the severity of depression, namely minimal, mild, moderate, and severe. We demonstrate the robustness of our model and advantages over state-of-the-art approaches. Finally, we conduct an extensive linguistic analysis and show common linguistic patterns between stress and depression.

Our main contributions can be summarized as follows:

\begin{itemize}
    \item We introduce a method, which injects linguistic features into transformer-based neural models.
    \item We perform model calibration by using label smoothing. We evaluate the calibration of our approaches by using two metrics. To the best of our knowledge, this is the first study exploiting label smoothing and utilizing calibration metrics.
    \item We contribute to the existing literature by performing a detailed linguistic analysis, which reveals significant differences in language between stressful/depressive and non-stressful/non-depressive posts.
\end{itemize}

\section{Related Work} \label{related_work}

\subsection{Stress Detection}

Existing research initiatives build on feature extraction and train of shallow ML algorithms. Muñoz and Iglesias \cite{MUNOZ2022103011} introduced three approaches for detecting psychological stress in social media text. In terms of the first approach, the authors extracted affective, social, syntactic, and topic-related features. Support Vector Machines  (SVMs), Logistic Regression (LR), and Stochastic Gradient Descent  (SGD) classifiers were trained. Regarding the second approach, the authors employed word embeddings, namely word2vec, GloVe, and fastText and trained the abovementioned machine learning classifiers. With regards to the third approach, the authors introduced an early fusion approach, where they concatenated the features and the word embeddings and trained the aforementioned classifiers. Guntuku et al. \cite{Chandra_Guntuku_Buffone_Jaidka_Eichstaedt_Ungar_2019} extracted LIWC, topic, TensiStrength, and engagements, i.e., time of posts, number of posts, number of posts between 12am-6am features and trained linear regression with regularization methods, including ridge, elastic-net, least absolute shrinkage and selection operator
(LASSO), and L2 penalized SVMs. They found that elastic-net showed marginally superior performance over the others. The authors proposed also domain adaptation methods, including EasyAdapt and Transfer Component Analysis.

Deep learning approaches and transformer-based models are being used for detecting stress. For instance, Turcan and McKeown \cite{turcan-mckeown-2019-dreaddit} introduced a dataset consisting of stressful and non-stressful text and applied machine learning techniques for differentiating stressful from non-stressful texts. Specifically, the authors extracted lexical, syntactic, social media features, word2vec, and BERT embeddings and trained various ML classifiers, including SVMs, LR, Naive Bayes, Perceptron, and Decision Trees  (DT). Also, deep neural networks were trained, including a two-layer bidirectional Gated Recurrent Neural Network (GRNN), CNN, and BERT. Yang et al. \cite{YANG2022102961} introduced a deep neural network to detect stress and depression in social media text. Specifically, the authors exploited MentalRoBERTa for obtaining token-level embeddings and commonsense transformers (COMET) for extracting mental state knowledge. Next, the authors exploited a GRU layer along with a scaled dot-product attention module. Finally, contrastive learning in a supervised manner was employed by the authors for making sentences with the same label cohesive and different labels mutually exclusive. Also, methods for increasing the training set have been applied. Specifically, Winata et al. \cite{8461990} proposed (Bi)LSTMs coupled with an Attention mechanism to classify psychological stress from self-conducted interview transcriptions. For expanding the size of the corpus, the authors applied distant supervision, where they automatically labelled tweets based on their hashtag content.

Hybrid models have also been proposed. For instance, Lin et al. \cite{7885098} experimented with capturing the users' social interactions in social media. Specifically, the authors proposed a deep neural network consisting of CNNs with cross autoencoders and a partially labeled factor graph. The authors exploited both tweet-level and user-level attributes. In terms of the tweet-level attributes, they exploited linguistic, i.e., positive and negative emotion words, punctuation marks, emoticons, and so on, visual, i.e., saturation, brightness, five-color theme, and so on, and social features, i.e., number of comments, retweets, likes. Regarding user-level attributes, they extracted features pertinent to their posting behaviour and social interaction.

Emotion-enhanced approaches in conjunction with explainability techniques have been introduced. Turcan et al. \cite{turcan-etal-2021-emotion} proposed three emotion-enhanced models that incorporate emotional information in various ways to enhance the task of binary stress prediction. In terms of the first approach, the authors exploited two single-task models sharing the same BERT representation layers. Specifically, the authors trained the stress task with the Dreaddit data and the emotion task with the GoEmotions or Vent data. Regarding the second model, the authors proposed a multi-task learning framework, where the two tasks were performed at the same time. With regards to the third model, the authors fine-tune first a BERT model using the emotion task and then apply this fine-tuned BERT model to stress detection. Finally, the authors introduced a new framework for model interpretation using  local interpretable model-agnostic explanations (LIME).

\subsection{Depression Detection}

Some studies have focused on the extraction of features and then the train of shallow machine learning classifiers. For instance, Tadesse et al. \cite{8681445} extracted n-grams via the tf-idf approach, LIWC features, and LDA topics. Then, they trained LR, SVM, Random Forest (RF), AdaBoost, and Multilayer Perceptron (MLP). Results showed that the bigram features trained on an SVM classifier achieved 80.00\% accuracy, while the best accuracy accounting for 91.00\% was achieved by exploiting the MLP classifier with all the features, i.e., LIWC, LDA, and bigrams. Liu and Shi \cite{10.3389/fpsyg.2021.802821} extracted a set of textual features, namely part-of-speech, emotional words, personal pronouns, polarity, and so on, and a set of features indicating the posting behaviour of the user, i.e., posting habits and time. Next, feature selection techniques were applied, including recursive elimination, mutual information, extreme random tree. Finally, naive bayes, k-nearest neighbor, regularized logistic regression, and support vector machine were used as base learners, and a simple logistic regression algorithm was used as a combination strategy to build a stacking model. Nguyen et al. \cite{6784326} extracted a set of features, including LDA topics, LIWC features, affective features by using the affective norms for english words (ANEW) lexicon, and mood labels. The authors trained a LASSO regression classifier for detecting depressive posts and analyzing the importance of each feature. The authors applied also statistical tests and found significant differences between depressive and non-depressive posts. Tsugawa et al. \cite{10.1145/2702123.2702280} extracted features and trained an SVM classifier to detect depression in Twitter. Specifically, the authors extracted the frequency of words used in tweets, ratio of tweet topics found by LDA, ratio of positive and negative words, and many more. Pirina and Çöltekin \cite{pirina-coltekin-2018-identifying} collected several corpora and trained an SVM classifier using character and word n-grams. Doc2vec and tf-idf features were extracted and given as input to AdaBoost, LR, RF, and SVM for identifying the severity of depression.

Recently, deep learning approaches have introduced, since they obtain better performance than the traditional ML algorithms and do not often require the tedious procedure of feature extraction. For example, Wani et al. \cite{9904737} represented words as word2vec and tf-idf approach and trained a deep neural network consisting of CNNs and LSTMs. Kim et al. \cite{kim2020deep} collected a dataset consisting of posts written by people, who suffer from mental disorders, including depression, anxiety, bipolar, borderline personality disorder, schizophrenia, and autism. This study developed six binary classification models for detecting mental disorders, i.e., depression vs. non-depression, and so on. Specifically, the authors utilized the tf-idf approach and trained an XGBoost classifier. Next, the authors used the word2vec and trained a CNN model.  Naseem et al. \cite{10.1145/3485447.3512128} reformulated depression identification as an ordinal classification problem, where they used four depression severity levels. The authors introduced a deep neural network consisting of Text Graph Convolutional Network, BiLSTM, and Attention layer.  A similar approach was proposed by Ghosh and Anwar \cite{9447025}, where the authors extracted features and trained LSTMs for estimating the depression intensity levels. A hybrid deep neural network consisting of CNN and BiLSTM was introduced by Kour and Gupta \cite{kour2022hybrid}. Zogan et al. \cite{10041797} introduced the first dataset including posts from users with and without depression during COVID-19 and presented a new hierarchical convolutional neural network. An emotion-based attention network model was proposed by Ren et al. \cite{info:doi/10.2196/28754}, where the authors extracted the positive and negative words and passed through two separate BiLSTM layers followed by Attention layers.

Ensemble strategies have also been explored in the literature. This means that multiple models are trained separately and the final decision is taken usually by a majority voting approach. For instance, an ensemble strategy was introduced by Ansari et al. \cite{9733425}. Firstly, the authors exploited some sentiment lexicons, including AFINN, NRC, SenticNet, and multi-perspective question answering (MPQA), extracted features, and applied Principal Component Analysis for reducing the dimensionality of the feature set. A Logistic Regression classifier was trained using the respective feature set. Next, the authors trained an LSTM neural network coupled with an attention mechanism. Finally, the authors combined the predictions of these two approaches via an ensemble method. Also, an ensemble approach was proposed by Trotzek et al. \cite{8580405}. Firstly, the authors trained a Logistic Regression classifier using as input user-level linguistic metadata. Specifically, the authors extracted LIWC features, length of the text, four readability scores, and so on. Next, the authors trained a CNN model. Finally, the authors combined the outputs of these approaches via a late fusion strategy, i.e., by averaging the predictions of the classifiers. Figuerêdo et al. \cite{FIGUEREDO2022100225} designed a CNN along with early and late fusion strategies. Specifically, the authors exploited fastText and GloVe embeddings. In the early fusion approach, multiple word embeddings were concatenated and passed to the CNN model. In the late fusion strategy, a majority-vote approach was performed based on the predictions of multiple CNN models. The CNN model comprised a simple convolution layer, max-pooling, fully connected layers, and Concatenated Rectified Linear Units as the activation function.

Explainable approaches have also been introduced. Souza et al. \cite{9723579} introduced a stacking ensemble neural network, which addresses a multilabel classification task. Specifically, the proposed architecture consists of two levels. In the first level, binary base classifiers were trained with two distinct roles, i.e, expert and differentiating. The expert base classifiers were used for differentiating between users belonging to the control group and those diagnosed with anxiety, depression, or comorbidity. The differentiating base models aimed at distinguishing between two target conditions, e.g., anxiety vs. depression. In the second level, a meta-classifier uses the base models’ outputs to learn a mapping function that manages the multi-label problem of assigning control or diagnosed labels. The authors used LSTMs and CNNs. Finally, this study explored  Shapley additive explanations (SHAP) metrics for identifying the influential classification features. Zogan et al. \cite{zogan2022explainable} proposed also an explainable approach, where textual, behavioural, temporal, and semantic aspect features from social media were exploited. An hierarchical attention network was used in terms of explainable purposes. An hierarchical attention network was also used by Uban et al. \cite{UBAN2021480}, where the authors extracted a feature set consisting of content, style, LIWC, and emotions/sentiment features. An interpretable approach was proposed by Song et al. \cite{song-etal-2018-feature}, where the authors introduced the Feature Attention Network. The Feature Attention Network consists of  four feature networks, each of which analyzes posts based on an
established theory related to depression and a post-level attention on top of the networks. However, this method did not attain satisfactory results.

Recently, transformer-based models have been applied in the task of depression detection in social media. Specifically, Boinepelli et al. \cite{boinepelli-etal-2022-leveraging} introduced a method for finding the subset of posts that would be a good representation of all the posts made by the user. Firstly, they employed BERT and computed the embeddings for all posts made by the user. Next, they used a clustering and ranking algorithm. After finding the representative posts per user, the authors added domain specific elements by exploiting RoBERTa. Finally, the authors experimented with two ways for diagnosing depression, i.e., by either employing a majority-vote approach or training a hierarchical attention network. Anantharaman et al. \cite{anantharaman-etal-2022-ssn} fine-tuned a BERT model for classifying the signs of depression into three labels namely “not depressed”, “moderately depressed”, and “severely depressed”. Similarly, Nilsson and Kovács \cite{nilsson-kovacs-2022-filipn} exploited a BERT model and used abstractive summarization techniques for data augmentation. Zogan et al. \cite{10.1145/3404835.3462938} presented an abstractive-extractive automatic text summarization model based on BERT, k-Means clustering, and  bidirectional auto-regressive transformers (BART). Then, they proposed a deep learning framework, which combines user behaviour and user post history or user activity.

Multimodal approaches combining both text and images have also been proposed. For instance, a multimodal approach was introduced by Ghosh et al. \cite{9567734} for detecting depression in Twitter. Specifically, the authors utilized the user's description and profile image. The authors used the IBM Watson NaturalLanguageUnderstanding tool and extracted sentiment and emotion information for all user descriptions along with the possible categories (at most 3) that the description may belong to. Next, the authors designed a neural network consisting of BiGRU, Attention layers, Convolution layers, and dense layers. The authors used GloVe embeddings. The proposed architecture can predict whether the user suffers from depression or not as well as predict the sadness, joy, fear, disgust, and anger score. Li et al. \cite{li2023mha} exploited text, pictures, and auxiliary information (post time, dictionary, social information) and used attention mechanism within and between the modalities at the same time. The authors exploited TextCNN, ResNet-18, and fully connected layers for extracting representation vectors of text, images, and auxiliary information respectively. A multimodal approach was proposed by Cheng and Chen \cite{cheng2022multimodal}, where the authors exploited texts, images, posting time, and the time interval between the posts in Instagram. Shen et al. \cite{ijcai2017p536} collected multimodal datasets and extracted six depression-oriented feature groups, namely social network, user profile, visual, emotional, topic-level, and domain-specific features. Gui et al. \cite{Gui_Zhu_Zhang_Peng_Zhou_Ding_Chen_2019} combined texts and images and  proposed a new cooperative multi-agent reinforcement learning method.

Multitask approaches have been introduced. A multitask approach was introduced by Zhou et al. \cite{ZHOU2021102119}. Specifically, the authors proposed a hierarchical attention network consisting of BiGRU layers and integrated LDA topics. The main task was the identification of depression, i.e., binary classification task, while the auxiliary task was the prediction of the domain category of the post, i.e., multiclass classification task. Both multitask and multimodal approaches were introduced by Wang et al. \cite{WANG2022727}. The authors extracted a total of ten features from text, social behaviour, and pictures. XLNet, BiGRU coupled with Attention layers, and Dense layers were used.

\subsection{Related Work Review Findings}
Existing research initiatives rely on the feature extraction process and the train of shallow machine classifiers targeting at diagnosing mental disorders in social media. This fact demands domain expertise and does not generalize well to new data. Other existing approaches train CNNs, BiLSTMs, or employ hybrid models and ensemble strategies. Recently, transformer-based models have been used also. Only few works have experimented with injecting linguistic, including emotion, features into deep neural networks. These approaches employ multi-task learning models, fine-tuning, or multimodal approaches. All these approaches employing transformer-based models usually fine-tune these models. None of these approaches have used modifications of BERT aiming to enhance its performance by injecting into it external knowledge. Also, no prior work has taken into account model calibration creating in this way overconfident models.

Therefore, our work differs from the existing research initiatives, since we \textit{(i)} present a new method, which injects linguistic features into transformer-based models, \textit{(ii)} apply label smoothing for calibrating our model and evaluate both the performance and calibration of our proposed models, and \textit{(iii)} present a methodology for gaining linguistic insight into the tasks of depression and stress investigating their common characteristics.

\section{Methodology} \label{methodology}

\subsection{Architecture}

\begin{figure*}[!htb]
\centering
         \includegraphics[width=\textwidth]{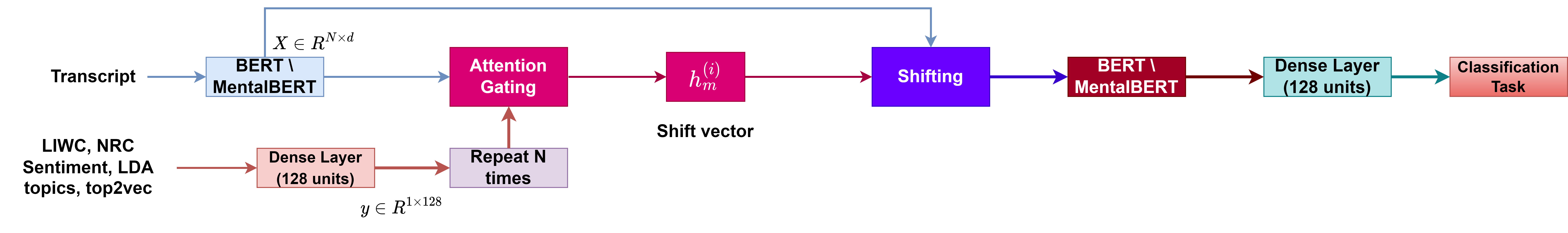}
         \caption{Our Proposed Architecture}
         \label{proposed_architecture}
\end{figure*}

In this section, we describe our proposed approach for detecting stressful and depressive posts in social media. Our proposed method is based on the work introduced by Rahman et al. \cite{rahman-etal-2020-integrating}, and Jin and Aletras \cite{jin-aletras-2020-complaint}. Instead of cross-modal interactions, we inject extra linguistic information as alternative views of the data into pretrained language models. Our proposed architecture is illustrated in Fig.~\ref{proposed_architecture}. 

Specifically, we use the following feature vectors:

\begin{itemize}
    \item \textbf{NRC.} The NRC Emotion Lexicon is a list of English words and their associations with eight basic emotions (anger, fear, anticipation, trust, surprise, sadness, joy, and disgust) and two sentiments (negative and positive) \cite{https://doi.org/10.1111/j.1467-8640.2012.00460.x}. Each text is represented as a 10-d vector, where each element is the proportion of tokens belonging to each category.
    \item \textbf{LIWC.} LIWC  is a dictionary-based approach to count words in linguistic, psychological, and topical categories \cite{pennebaker2001linguistic}. We use LIWC 2022 \cite{boyd2022development} to represent each text as a 117-d vector.
    \item \textbf{LDA topics.} Before training the LDA model, we remove stop words and punctuation. We exploit LDA (with 25 topics) and extract 25 topic probabilities per text \cite{10.5555/944919.944937}. These probabilities describe the topics of interest of each text. Inspired by Liu et al. \cite{liu-etal-2016-detecting}, we use the following feature vector:
    \begin{itemize}
        \item \textbf{Global Outlier Standard Score (GOSS):} For evaluating the $i^{th}$ text's interest on a certain topic $k$, compared to the rest of the texts, we use the GOSS feature:
        \begin{equation}
            \mu (x_k) = \frac{\sum_{i=1}^{n} x_{ik}}{n}
        \end{equation}
        \begin{equation}
            GOSS(x_{ik})=\frac{x_{ik}-\mu (x_k)}{\sqrt{\sum_{i} \left(x_{ik}-\mu \left(x_k \right)\right)^2}}
        \end{equation}
    \end{itemize}
    Therefore, each text is represented as a 25-d vector.
    \item \textbf{Top2Vec:} Top2Vec \cite{angelov2020top2vec} is an algorithm for topic modelling, which automatically detects topics present in text and generates jointly embedded topic, document and word vectors. After training Top2Vec by exploiting the Universal Sentence Encoder, each text is represented as a 512-d vector.
\end{itemize}

We experiment with the following pretrained models: BERT \cite{devlin-etal-2019-bert} and MentalBERT \cite{ji-etal-2022-mentalbert}.

First, we pass each text through the aforementioned transformer-based models. Let $C \in \mathcal{R}^{N \times d}$ be the output of the transformer-based models, where $N$ denotes the sequence length, while $d$ denotes the dimensionality of the models. We have omitted the dimension corresponding to the batch size for the sake of simplicity.

Then, we project the feature vectors to dimensionality equal to $128$. We repeat the feature vector $N$ times, so as to ensure that the feature vector and the output of the transformer-based models can be concatenated. Given the word representation $e^{(i)}$, we concatenate $e^{(i)}$ with feature vectors, i.e., $h_v ^{(i)}$.

\begin{equation}
    w^{(i)} _v = \sigma \left(W_{hv} [e^{(i)};h^{(i)} _v] + b_v \right)
\end{equation}

where $\sigma$ denotes the sigmoid activation function, $W_{hv}$ is a weight matrix, and $w^{(i)} _v$ corresponds to the gate. $b_v$ is the scalar bias.

Next, we calculate a shift vector $h^{(i)} _m$ by multiplying the embeddings with the gate.

\begin{equation}
    h^{(i)} _m = w^{(i)} _v \cdot \left(W_v h^{(i)} _v \right) + b^{(i)} _m 
\end{equation}

where $W_v$ is a weight matrix and $b^{(i)} _m$ is the bias vector.

Next, we apply the Multimodal Shifting component aiming to dynamically shift the word representations by integrating the shift vector $h^{(i)} _m$ into the original word embedding.

\begin{equation}
    e^{(i)} _m = e^{(i)} + \alpha h^{(i)} _m
    \label{equationmulti}
\end{equation}

\begin{equation}
    \alpha = min \left(\frac{||e^{(i)}||_2}{||h^{(i)} _m||_2}\beta,1 \right)
    \label{beta_model}
\end{equation}
, where $\beta$ is a hyperparameter. Then, we apply a layer normalization \cite{ba2016layer} and dropout layer \cite{JMLR:v15:srivastava14a} to $e^{(i)} _m$. Next, the combined embeddings are fed to a BERT/MentalBERT model. 

We get the classification [CLS] token of this model and pass it through a Dense layer consisting of 128 units with a ReLU activation function. Finally, we use a dense layer consisting of either two units (binary classification task) or four units (multiclass classification task).

We denote our proposed models as Multimodal BERT (M-BERT) and Multimodal MentalBERT (M-MentalBERT) followed by the linguistic features which are integrated into them. For example, the injection of LIWC features into a BERT model is denoted as M-BERT (LIWC).

\subsection{Model Calibration}

To prevent the model becoming too overconfident, we use label smoothing \cite{7780677,NEURIPS2019_f1748d6b}. Specifically, label smoothing calibrates learned models so that the confidences of their predictions are more  aligned with the accuracies of their predictions. 

For a network trained with hard targets, the cross-entropy loss is minimized between the true targets $y_k$ and the network's outputs $p_k$, as in $H(y,p)=\sum_{k=1}^K -y_k log(p_k)$, where $y_k$ is "1" for the correct class and "0" for the other. For a network trained with label smoothing, we minimize instead the cross-entropy between the modified targets $y_k ^{LS_u}$ and the network's outputs $p_k$.

\begin{equation}
    y_k ^{LS_u} = y_k \cdot (1-\alpha)+ \frac{\alpha}{K}
    \label{label_smoothing}
\end{equation}

\begin{equation}
    H(y,p)=\sum_{k=1}^{K} -y_k ^{LS_u} \cdot \log \left(p_k\right)
\end{equation}

, where $\alpha$ is the smoothing parameter and $K$ is the number of classes.

\section{Experiments}

\subsection{Datasets}

\noindent \textbf{Dreaddit.} This dataset includes stressful and non-stressful texts posted by users of Reddit \cite{turcan-mckeown-2019-dreaddit}. Specifically, these posts belong to five domains, namely abuse, anxiety, financial, social, and Post-Traumatic Stress Disorder (PTSD). A subset of data has been annotated using Amazon Mechanical Turk.  This dataset includes also lexical, syntactic, and social media features per post. The dataset has been divided by the authors into a train and a test set. The train set comprises 1488 stressful texts and 1350 non-stressful ones, while the test set includes 369 stressful texts and 346 non-stressful ones.

\noindent \textbf{Depression\_Mixed.} This dataset \cite{pirina-coltekin-2018-identifying} consists of 1482 non-depressive posts and 1340 depressive posts. These posts have been written by users in Reddit and English depression forums.

\noindent \textbf{Depression\_Severity.} This dataset includes posts in Reddit \cite{10.1145/3485447.3512128} and assigns each post to a severity level, i.e., minimal (2587 posts), mild (290 posts), moderate (394 posts), and severe form of depression (282 posts).


\subsection{Experimental Setup}

We use the Adam optimizer with a learning rate of 0.001. We apply \textit{StepLR} with a step size of 5 and a gamma of 0.1. We use batch size of 8. With regards to \textit{Depression\_Mixed} dataset, we split the dataset into a train and a test set ($80\%-20\%$) similar to Ansari et al. \cite{9733425}. In terms of \textit{Dreaddit} dataset, we use the test set provided by Turcan and McKeown \cite{turcan-mckeown-2019-dreaddit}. Regarding \textit{Depression\_Severity} dataset, we use 5-fold stratified cross-validation, since the study \cite{10.1145/3485447.3512128} has also exploited cross-validation. All train sets are divided into a train and a validation set. Regarding \textit{Depression\_Severity} dataset, we apply \textit{EarlyStopping} with a patience of 7 epochs based on the validation loss. In terms of \textit{Depression\_Mixed} and \textit{Dreaddit} dataset, we train our introduced model for a maximum of 30 epochs, choose the epoch with the smallest validation loss, and test the model on the test set. We set $\beta$ of Eq.~\ref{beta_model} equal to 0.0001.\footnote{We experimented with values of $\beta$, including 0.01 and 0.001, but setting $\beta$ equal to 0.0001 yielded the best results.} We choose $\alpha$ of Eq.~\ref{label_smoothing} equal to 0.001. We use the Python library, namely Transformers \cite{wolf-etal-2020-transformers}, for BERT and MentalBERT. Specifically, we use the BERT base uncased version and the MentalBERT base uncased version. We use PyTorch \cite{NEURIPS2019_bdbca288} for performing our experiments. All experiments are trained on a single Tesla P100-PCIE-16GB GPU. 

\subsection{Evaluation Metrics}

\subsubsection{Performance}
In terms of the binary classification tasks, i.e., 0 for non-stressful and 1 for stressful texts or 0 for non-depressive and 1 for depressive texts, we use Precision, Recall, F1-score, and Accuracy to evaluate the performance of our proposed approach.  We use these metrics similar to Wani et al. \cite{9904737}.

Regarding multiclass classification task reported on Depression\_Severity dataset, we use Weighted Precision, Weighted Recall, and Weighted F1-score. We use these metrics similar to Mishra et al. \cite{mishra-etal-2018-author}.

\subsubsection{Calibration}
We evaluate the calibration of our model using the metrics proposed by relevant literature \cite{naeini2015obtaining,Nixon_2019_CVPR_Workshops,pmlr-v70-guo17a}. Specifically, we use the metrics mentioned below:

\begin{itemize}
    \item \textbf{Expected Calibration Error (ECE).} The calibration error is the difference between the fraction of predictions in the bin that are correct (accuracy) and the mean of the probabilities in the bin (confidence). First, we divide the predictions into $M$ equally spaced bins (size $1/M$).

    \begin{equation}
        acc(B_m) = \frac{1}{|B_m|}\sum_{i \in B_m} 1(\hat{y}_i = y_i)
    \end{equation}
    
    \begin{equation}
        conf(B_m) = \frac{1}{|B_m|}\sum_{i \in B_m} \hat{p}_i
    \end{equation}
    , where $y_i$ and $\hat{y}_i$ are the true and predicted labels for the sample $i$ and $\hat{p}_i$ is the confidence (predicted probability value) for sample $i$.
    
    \begin{equation}
        ECE = \sum_{m=1}^{M} \frac{|B_m|}{N} \left|acc(B_m)-conf(B_m)\right|
    \end{equation}
    , where $N$ is the total number of data points and $B_m$ is the group of samples whose predicted probability values falls into the interval $I_m=\left(\frac{m-1}{M},\frac{m}{M}\right]$.

    Perfectly calibrated models have an ECE of 0.
    
    \item \textbf{Adaptive Calibration Error (ACE).} Adaptive Calibration Error uses an adaptive scheme which spaces the bin intervals so that each contains an equal number of predictions.

    \begin{equation}
        ACE=\frac{1}{KR} \sum_{k=1}^{K} \sum_{r=1}^{R} |acc(r,k)-conf(r,k)|
    \end{equation}
    , where $acc(r, k)$ and $conf(r, k)$ are the accuracy and confidence of adaptive calibration range $r$ for class label $k$, respectively; and $N$ is the total number of data points. Calibration range $r$ is defined by the [$N/R$]th index of the sorted and thresholded predictions.
\end{itemize}

\subsection{Baselines}

We use the following baselines as comparisons with our proposed approaches:

\begin{itemize}
    \item \textbf{BERT, MentalBERT:} We fine-tune these pretrained language models in order to explore whether our method of injecting linguistic information to pretrained models leads to performance improvement.

    In terms of Depression\_Mixed dataset, we report the performance of BERT obtained by Yang et al. \cite{YANG2022102961}. We finetune MentalBERT and report its performance on this dataset.

    With regards to the Dreaddit dataset, we report the performance of BERT and MentalBERT obtained by Turcan and McKeown \cite{turcan-mckeown-2019-dreaddit}, and Ji et al. \cite{ji-etal-2022-mentalbert} respectively.

    Regarding Depression\_Severity dataset, we finetune BERT and MentalBERT and report their performances. 

    We do not report calibration metrics for these models, since our goal in this case is to compare only the performances of these models with our proposed approaches.
    \item \textbf{Proposed Approaches (without label smoothing):} We train the proposed models introduced in Section~\ref{methodology} without label smoothing. We explore whether label smoothing leads to performance improvement and better calibration of our models.
\end{itemize}

\section{Results}

The results of our proposed approach are reported in Tables~\ref{Results_1} and \ref{Results_3}. Specifically, Table~\ref{Results_1} reports the performances of our proposed approaches on the Depression\_Mixed and Dreaddit datasets, while Table~\ref{Results_3} reports the results on the Depression\_Severity dataset.

\begin{table*}[!htb]
\tiny
\centering
\caption{Performance comparison among proposed models and baselines using the DEPRESSION\_MIXED and DREADDIT datasets.}
\begin{tabular}{lcccc|cc||cccc|cc}
\toprule
\multicolumn{1}{c}{}&\multicolumn{6}{c||}{\textbf{Depression\_Mixed}}&\multicolumn{6}{c}{\textbf{Dreaddit}}\\ \midrule
\textbf{Model} & Prec. & Rec. & F1-score & Acc. & ECE & ACE & Prec. & Rec. & F1-score & Acc. & ECE & ACE \\
\midrule
\multicolumn{13}{>{\columncolor[gray]{.8}}l}{\textbf{Baselines}} \\
BERT & 91.40 & 91.40 & 91.40 & - & - & - & 75.18 & 86.99 & 80.65 & - &-&-\\
MentalBERT & 89.27 & 93.14 & 91.17 & 91.15 & - & - & - & 80.28 & 80.04 & - & - & - \\
\midrule
\multicolumn{13}{>{\columncolor[gray]{.8}}l}{\textbf{Baselines - Proposed Approaches (without label smoothing)}} \\
M-BERT (NRC) & 90.56 & 91.84 & 91.20 & 91.15 & 0.072 & 0.081 & 76.70 & 85.64 & 80.92 & 79.16 & 0.117 & 0.125 \\
M-BERT (LIWC) & 90.98 & 92.02 & 91.49 & 92.04 & 0.054 & 0.055 & 79.74& 84.28&81.95 & 80.84& 0.118& 0.110\\
M-BERT (LDA topics) & 88.07 & 95.80 & 91.77 & 92.04 & 0.071& 0.071 & 76.87& 86.68& 81.48& 79.64& 0.103& 0.108\\
M-BERT (top2vec) & 90.97 & 92.99 & 91.97 & 92.21 & 0.057 & 0.069 & 78.25& 84.82& 81.40& 80.00& 0.075& 0.133\\
M-MentalBERT (NRC) & 90.65 & 92.65 & 91.64 & 91.86 & 0.031 & 0.054 & 80.31 & 82.93 & 81.60 & 80.69& 0.099& 0.101\\
M-MentalBERT (LIWC) & 93.49 & 87.78 & 90.55 & 91.50 & 0.057 & 0.056 & 79.85 & 85.91 & 82.77 & 81.53&0.158& 0.153\\
M-MentalBERT (LDA topics) & 87.97 & 93.09 & 90.46 & 90.44 & 0.089 & 0.086 & 78.16 & 85.60& 81.71& 80.19& 0.109& 0.134\\
M-MentalBERT (top2vec) & 91.63 & 93.77& 92.69& 93.27& 0.058& 0.054&79.70 & 85.09& 82.31& 81.12&0.123 & 0.149\\
\midrule
\multicolumn{13}{>{\columncolor[gray]{.8}}l}{\textbf{Proposed Approaches (with label smoothing)}} \\
M-BERT (NRC) & 89.82 & 94.81 & 92.25 & 92.39 & 0.059 & 0.065 & 76.51& 89.16& 82.35& 80.28& 0.094& 0.105\\
M-BERT (LIWC) & 93.06 & 91.78 & 92.41 & 92.21 & 0.034 & 0.044 & 77.21& 89.97& 83.10& 81.12& 0.114& 0.125\\
M-BERT (LDA topics) & 90.16 & 92.71& 91.42& 92.39& 0.063& 0.067& 80.10& 84.24& 82.12& 81.04& 0.063& 0.080\\
M-BERT (top2vec) & 90.34 & 94.93 & 92.58 & 92.57 & 0.049 & 0.056 & 78.41 & 85.64& 81.87& 80.42& 0.066& 0.083\\
M-MentalBERT (NRC) & 91.44 & 92.52 & 91.98 & 92.74 & 0.042 & 0.057 & 79.70& 86.18& 82.81& 81.53& 0.082& 0.101\\
M-MentalBERT (LIWC) & 94.96 & 89.42 & 92.11 & 92.57 &  0.055 & 0.057 & 79.41& 87.80& 83.40& 81.96& 0.116& 0.112\\
M-MentalBERT (LDA topics) & 94.81 & 90.78 & 92.75 & 92.92 & 0.047 & 0.049 & 79.11 & 87.50&83.10 &81.60 &0.096 &0.111 \\
M-MentalBERT (top2vec) & 96.12 & 90.18 & 93.06 & 93.45 & 0.033 & 0.043 & 78.92& 87.26&82.88 &81.39 & 0.121 & 0.137\\
\bottomrule
\end{tabular}
\label{Results_1}
\end{table*}

\begin{table}[!htb]
\tiny
\centering
\caption{Performance comparison among proposed models and baselines using the DEPRESSION\_SEVERITY dataset.}
\begin{tabular}{lccc|cc}
\toprule
\textbf{Model} & W. Prec. & W. Rec. & W. F1-score & ECE & ACE \\
\midrule
\multicolumn{6}{>{\columncolor[gray]{.8}}l}{\textbf{Baselines}} \\
BERT & 72.99 & 71.97 & 71.00 & - & - \\
MentalBERT & 73.35 & 70.81 & 71.67 & - & -  \\
\midrule
\multicolumn{6}{>{\columncolor[gray]{.8}}l}{\textbf{Baselines - Proposed Approaches (without label smoothing)}} \\
M-BERT (NRC) & 74.48 & 70.08 & 69.96 & 0.107 & 0.076 \\
M-BERT (LIWC) & 73.77 & 71.74 & 72.13 & 0.110 & 0.078\\
M-BERT (LDA topics) & 74.25 & 71.80 & 71.28 & 0.114 & 0.079  \\
M-BERT (top2vec) & 72.93 & 71.97 & 71.00 & 0.086 & 0.071  \\
M-MentalBERT (NRC) & 74.43 & 72.58 & 69.96 & 0.097 & 0.069 \\
M-MentalBERT (LIWC) & 72.39 & 72.53 & 71.95 & 0.112 & 0.075  \\
M-MentalBERT (LDA topics) & 73.83 & 72.58 & 72.58 & 0.118 & 0.078  \\
M-MentalBERT (top2vec) & 74.63 & 72.39 & 72.06 & 0.103 & 0.075  \\
\midrule
\multicolumn{6}{>{\columncolor[gray]{.8}}l}{\textbf{Proposed Approaches (with label smoothing)}} \\
M-BERT (NRC) &74.04 & 72.84 & 72.81 & 0.102 & 0.074  \\
M-BERT (LIWC) & 73.68 & 72.16 & 72.37 & 0.094 & 0.069  \\
M-BERT (LDA topics) & 73.24 & 71.46 & 71.42 & 0.112 & 0.078   \\
M-BERT (top2vec) & 73.36 & 72.64 & 72.30 & 0.113 & 0.074  \\
M-MentalBERT (NRC) & 73.03 & 71.23 & 71.46 & 0.112 & 0.079  \\
M-MentalBERT (LIWC) & 73.21 & 73.15 & 72.43 & 0.099 & 0.071   \\
M-MentalBERT (LDA topics) & 73.74 & 73.23 & 73.16 & 0.111 & 0.075   \\
M-MentalBERT (top2vec) & 73.68 & 72.70 & 72.67 & 0.094 & 0.071   \\
\bottomrule
\end{tabular}
\label{Results_3}
\end{table}

In terms of the Dreaddit dataset, we first compare our proposed approaches without label smoothing with the BERT and MentalBERT models. First, we observe that the integration of linguistic features into the BERT model improves the performance obtained by BERT. Specifically, M-BERT (LIWC) yields the highest F1-score accounting for 81.95\% surpassing BERT by 1.30\%. At the same time M-BERT (LIWC) outperforms M-BERT (NRC),  M-BERT (LDA topics), and M-BERT (top2vec) in Precision, F1-score, and Accuracy. Similarly, the injection of LIWC features into the MentalBERT model yields an F1-score of 82.77\% outperforming MentalBERT by 2.73\% and the other approaches by 0.46-1.17\%. M-MentalBERT (NRC) yields the lowest F1-score accounting for 81.60\%. With regards to the proposed approaches with label smoothing, we observe that they outperform the proposed approaches without label smoothing in terms of both performance and calibration metrics. Specifically, we observe that M-BERT (LIWC) with label smoothing attains the highest F1-score and Accuracy accounting for 83.10\% and 81.12\% respectively outperforming M-BERT (LIWC) without label smoothing in F1-score by 1.15\% and in Accuracy by 0.28\%. The integration of LIWC features into the MentalBERT model with label smoothing obtains better performance than MentalBERT in F1-score by 3.36\% and better performance than M-MentalBERT (LIWC) without label smoothing in F1-score by 0.63\%. In addition, we observe that label smoothing leads to improvement in calibration metrics. For instance, M-BERT (NRC) with label smoothing improves both ECE and ACE by 0.023 and 0.020 respectively in comparison with M-BERT (NRC) without label smoothing.

Regarding the Depression\_Mixed dataset, we first compare our proposed approaches without label smoothing with the BERT and MentalBERT models. We observe that the injection of linguistic features, except for NRC features, into the BERT model improves the F1-score. Specifically, we observe that the injection of top2vec features yields the highest F1-score and Accuracy accounting for 91.97\% and 92.21\% respectively, surpassing the performance of the BERT model in F1-score by 0.57\%. We speculate that the injection of top2vec features obtains better performance than the injection of features derived by LDA topics, i.e., GOSS features, since the top2vec algorithm is capable of identifying the number of topics automatically. In terms of MentalBERT, we observe that the injection of top2vec features obtains an F1-score of 92.69\% surpassing MentalBERT by 1.52\%. We observe that the integration of NRC and top2vec features improves the performance obtained by MentalBERT. Regarding the proposed approaches with label smoothing, we observe that these models attain better performances than the ones obtained by the models without label smoothing. Specifically, we observe that M-BERT (top2vec) with label smoothing surpasses the respective model without label smoothing in F1-score and Accuracy by 0.61\% and 0.36\% respectively. Similarly, M-MentalBERT (top2vec) with label smoothing obtains the highest F1-score and Accuracy accounting for 93.06\% and 93.45\% respectively. This model surpasses the respective model without label smoothing in F1-score and Accuracy by 0.37\% and 0.18\%. Except for the improvement of the performance metrics, i.e., Precision, Recall, F1-score, and Accuracy, we observe that the models with label smoothing obtain better results in terms of the calibration metrics, i.e., ECE and ACE, than the ones obtained by the models without label smoothing. For example, we observe that M-BERT (top2vec) with label smoothing improves the ECE and ACE scores obtained by M-BERT (top2vec) without label smoothing by 0.008 and 0.013 respectively. Similarly, M-MentalBERT (LDA topics) with label smoothing improves the ECE and ACE scores obtained by M-MentalBERT (LDA topics) without label smoothing by 0.042 and 0.043 respectively.

With regards with the Depression\_Severity dataset, we first compare our proposed approaches without label smoothing with the BERT and MentalBERT models. We observe that the integration of LIWC features and features extracted by LDA topic modelling, i.e., GOSS features, into the BERT model leads to a performance surge in comparison with the BERT model. Specifically, M-BERT (LIWC) outperforms BERT in weighted F1-score by 1.13\%. At the same time, the integration of all the features, except NRC, to a MentalBERT model yields to a performance improvement compared to the MentalBERT model. Specifically, M-MentalBERT (LDA topics) attains the highest weighted F1-score accounting for 72.58\% surpassing MentalBERT by 0.91\%. When it comes to proposed models with label smoothing, we observe an improvement in both the performance metrics and calibration ones. More specifically, the integration of NRC features to a BERT model obtains a weighted F1-score of 72.81\% outpeforming BERT by 1.81\%, M-BERT (NRC) without label smoothing by 2.85\%, and M-BERT (LIWC) without label smoothing by 0.68\%. In addition, M-MentalBERT (LDA topics)
with label smoothing obtains the highest F1-score accounting for 73.16\% surpassing MentalBERT by 1.49\% and M-MentalBERT (LDA topics) without label smoothing by 0.58\%. In terms of the calibration metrics, we observe that both ECE and ACE scores are improved when we apply label smoothing. For example, M-BERT (LIWC) with label smoothing obtains an ECE score of 0.094 and an ACE score of 0.069, which are improved by 0.016 and 0.009 respectively compared to the respective model without label smoothing.

\section{Linguistic Analysis}

\begin{table*}[!htb]
\tiny
\centering
\caption{LIWC Features associated with depressive/stressful and non-depressive/non-stressful posts, sorted by point-biserial correlation. All correlations are significant at \textit{p} < 0.05 after Benjamini-Hochberg correction.}
\begin{tabular}{|c|c||c|c|||c|c||c|c|}
\hline
\multicolumn{4}{|c|||}{\textbf{Depression\_Mixed}}&\multicolumn{4}{c|}{\textbf{Dreaddit}} \\ \hline
\multicolumn{2}{|c||}{\textbf{Non-Depressive}}&\multicolumn{2}{c|||}{\textbf{Depressive}}&\multicolumn{2}{c|}{\textbf{Non-Stressful}}&\multicolumn{2}{c|}{\textbf{Stressful}}\\
\hline \hline
\textbf{LIWC} & \textbf{corr.} & \textbf{LIWC} & \textbf{corr.} & \textbf{LIWC} & \textbf{corr.} & \textbf{LIWC} & \textbf{corr.}\\ \hline
Tone & 0.3156 & health & 0.4108 & Tone & 0.4232 & 1st person singular & 0.3975\\ \hline
Clout & 0.3022 & mental health & 0.3674 & clout & 0.3501 & tone\_neg & 0.3807\\ \hline
Social referents & 0.2914 & physical & 0.3603 & tone\_pos & 0.2574 & emo\_neg & 0.3434\\ \hline
shehe & 0.2634 & emo\_sad & 0.3506 & analytic & 0.2269 & dic & 0.3122\\ \hline
we & 0.2415 & tone\_neg & 0.3274& social & 0.2084 & authentic & 0.2964\\ \hline
social & 0.2401 & 1st person singular & 0.2974& you & 0.2082 & linguistic & 0.2955\\ \hline
male references & 0.2199 & Authentic & 0.2961& Social referents & 0.1890 & function & 0.2928\\ \hline
affiliation & 0.1960 & cognition & 0.2957& emo\_pos & 0.1630 & emo\_anx & 0.2409\\ \hline
female references & 0.1923 & emo\_neg & 0.2843& affiliation & 0.1495 & emotion & 0.2354\\ \hline
conversation & 0.1794 & cognitive processes & 0.2601& article & 0.1338 & pronoun & 0.2272\\ \hline
netspeak & 0.1741 & feeling & 0.2507& we & 0.1334 & feeling & 0.2162\\ \hline
culture & 0.1732 & focuspresent & 0.2139& prosocial & 0.1290 & personal pronouns & 0.2150\\ \hline
allpunc & 0.1667 & insight & 0.2138& number & 0.1284 & focuspresent & 0.2110\\ \hline
family & 0.1589 & emotion & 0.2120& OtherP & 0.1231 & adverb & 0.1925\\ \hline
technology & 0.1524 & negations & 0.2116& Social behavior & 0.1229 & negations & 0.1900\\ \hline
exclam & 0.1519 & verb & 0.2076& drives & 0.1130 & verb & 0.1862\\ \hline
analytic & 0.1507 & linguistic & 0.1893& polite & 0.1121 & apostro & 0.1854\\ \hline
period & 0.1458 & death & 0.1842& comma & 0.1079 & affect & 0.1811\\ \hline
drives & 0.1168 & function & 0.1834& curiosity & 0.0990 & conjunctions & 0.1630\\ \hline
OtherP & 0.1156 & all-or-none & 0.1748& reward & 0.0971 & auxverb & 0.1626\\ \hline
number & 0.1075 & dic & 0.1708& female & 0.0909 & WC & 0.1502\\ \hline
assent & 0.1013 & affect & 0.1609& determiners & 0.0896 & emo\_sad & 0.1482\\ \hline
tone\_pos & 0.0998 & adverb & 0.1588& visual & 0.0867 & swear & 0.1424\\ \hline
leisure & 0.0980 & illness & 0.1584& exclam & 0.0839 & cognition & 0.1325\\ \hline
Social behavior & 0.0937 & emo\_anx & 0.1458& culture & 0.0782 & emo\_anger & 0.1260\\ \hline
communication & 0.0920 & auxverb & 0.1442& lifestyle & 0.0777 & causation & 0.1188\\ \hline
lifestyle & 0.0917 & discrepancy & 0.1373& money & 0.0750 & allure & 0.1188\\ \hline
friend & 0.0812 & apostro & 0.1256& attention & 0.0709 & risk & 0.1099\\ \hline
curiosity & 0.0792 & want & 0.1146& allpunc & 0.0672 & cognitive processes & 0.1081\\ \hline
you & 0.0774 & achieve & 0.1136& leisure & 0.0635 & health & 0.1013\\ \hline
determiners & 0.0758 & pronoun & 0.1092& they & 0.0590 & all-or-none & 0.1008\\ \hline
politic & 0.0751 & lack & 0.1054& technology & 0.0574 & insight & 0.0988\\ \hline
relig & 0.0673 & differ & 0.1004& food & 0.0561 & conflict & 0.0984\\ \hline
focusfuture & 0.0666 & prepositions & 0.0951& politic & 0.0547 & physical & 0.0947\\ \hline
visual & 0.0660 & risk & 0.0928& communication & 0.0532 & illness & 0.0935\\ \hline
motion & 0.0647 & allure & 0.0923& quantity & 0.0528 & words per sentence (WPS) & 0.0904\\ \hline
money & 0.0626 & causation & 0.0916& BigWords & 0.0474 & need & 0.0897\\ \hline
ethnicity & 0.0591 & tentative & 0.0880& wellness & 0.0471 & mental health & 0.0893\\ \hline
article & 0.0493 &  time & 0.0811& work & 0.0465 & death & 0.0849\\ \hline
emo\_pos & 0.0480 & personal pronouns & 0.0801& religion & 0.0432 & impersonal pronoun & 0.0813\\ \hline
home & 0.0420 & impersonal pronoun & 0.0789& shehe & 0.0389 & fatigue & 0.0768\\ \hline
food & 0.0401 & perception & 0.0747& ethnicity & 0.0387 & memory & 0.0746\\ \hline
words per sentence (WPS) & 0.0391 & swear & 0.0697& - & - & time & 0.0726\\ \hline
Nonfluencies & 0.0385 & substances & 0.0686& - & - & certitude & 0.0557\\ \hline
- & - & memory & 0.0673& - & - & preception & 0.0397\\ \hline
- & - & BigWords & 0.0588& - & - & motion & 0.0385\\ \hline
- & - & adj & 0.0575& - &-& differ & 0.0374\\ \hline
- & - & certitude & 0.0547& - & - & - & -\\ \hline
- & - & wellness & 0.0457& - & - & - & -\\ \hline
- & - & moral & 0.0433& - & - & - & -\\ \hline
- & - & conflict & 0.0411& - & - & - & -\\ \hline
- & - & acquire & 0.0394& - & - &- & -\\ \hline
- & - & QMark & 0.0384& - & - & - & -\\ 
\hline
\end{tabular}
\label{correlations_depression}
\end{table*}

We finally perform an analysis on the Depression\_Mixed and Dreaddit datasets to uncover the peculiarities of stress and depression. Specifically, we seek to find the correlations of LIWC features with stressful/depressive and non-stressful/non-depressive texts. To do this, we adopt the methodology by Ilias and Askounis \cite{9769980}. First, we normalize LIWC features, so as to ensure that they sum up to 1 across each post. Next, we use the point-biserial correlation between each LIWC category and the label of the post. The output of the point-biserial correlation is a number ranging from -1 to 1. Positive correlations mean that the specific LIWC category is correlated with the stressful/depressive class (label 1), while negative correlations mean that the specific LIWC category is correlated with the non-stressful/non-depressive class (label 0).  We consider the absolute values of the correlations. Results are reported in Table~\ref{correlations_depression}. All the correlations are significant at $p < 0.05$ with Benjamini-Hochberg correction \cite{benjamini1995controlling} for multiple comparisons.

In terms of the \textit{Depression\_Mixed} dataset, we observe that the control group tends to use words with positive tone and emotion, i.e., good, well, happy, hope, and so on. In addition, healthy control group discusses topics of the everyday life, including lifestyle (work, home, school), culture (car, phone), politics (govern, congress), family, and friends (boyfriend, girlfriend, dude). Also, these people make plans for the future, thus use words indicating focus on the future (correlation equal to 0.0666).  However, it must be noted that this is a very weak correlation. On the other hand, people with depression focus on the present and do not make plans for the future. They discuss about negative topics, including death, illnesses, mental health, and substances. This can be justified by the fact that people with depression often have tendencies to suicide and believe that they cannot achieve anything. In addition, they use swear words, i.e., shit, fuck, damn, since they think that everything goes wrong in their life. Also, their posts are full of sadness, anxiety, and negative tone. 

Regarding \textit{Dreaddit} dataset, we observe similar patterns. Specifically, non-stressful posts include words with positive tone and emotion. People in non-stressful conditions discuss about topics pertinent to lifestyle, money, leisure, technology, food, work, religion, and many more. Also, people use personal pronouns, including 3rd person singular, 2nd person, and 1st person plural. Also, their posts include words indicating politeness. On the contrary, people in stressful conditions use swear words or words indicating interpersonal conflict, including fight, kill, attacking, and so on. The first person singular constitutes the LIWC category with the highest degree of correlation with the stressful class.  This result agrees with previous work \cite{PENNEBAKER2002271}, where it is mentioned that self-references by individuals present a surge in emotionally vulnerable conditions. Therefore, the use of first person singular indicates increased self-focus. In addition, similar to the depressive posts, we observe that stressful posts include words with negative emotion, anger, sadness, and anxiety. Also, topics discussed by these users are pertinent to illness, health, and death.

\section{Discussion}

Our study contributes to the literature by introducing the first approach of integrating extra linguistic information into pretrained language models based on transformers, namely BERT and MentalBERT. Specifically, we adapt M-BERT \cite{rahman-etal-2020-integrating} by replacing multimodal information with linguistic information. To be more precise, we extract NRC, LIWC, features derived by LDA topics, and top2vec features. We apply a Multimodal Adaptation Gate and exploit also a Shifting component for creating new combined embeddings which are given as input to BERT (and MentalBERT) models. In addition, motivated by the fact that in real-world decision making systems, classification networks must not only be accurate, but also should indicate
when they are likely to be incorrect, we apply label smoothing and evaluate our proposed approaches both in terms of classification and calibration. 

Therefore, our study is different from the state-of-the-art approaches described in Section~\ref{related_work}, since:
\begin{itemize}
    \item Prior works having proposed multimodal, multitask, ensemble strategies in conjunction with transformer-based models, have just fine-tuned these pretrained transformer-based models instead of using some modifications of them. Thus, this study is the first attempt to inject extra knowledge into BERT (and MentalBERT), in order to enhance its performance.
    \item All the prior works evaluate only the classification performance of their approaches neglecting the confidence of the prediction. To tackle this, this is the first study in the task of stress and depression detection through social media posts utilizing label smoothing and evaluating both the classification performance and the calibration of the models. 
    \item Finally, this is the first study utilizing features derived by LDA topics, namely the Global Outlier Standard Score, which captures the text's interest compared to other texts.
\end{itemize}

From the results of this study, we found that:
\begin{itemize}
    \item \textit{Finding 1:} The integration of linguistic features into transformer-based models yields to an increase in the classification performance. However, it is worth noting that in some cases this improvement is limited. For instance, the integration of LIWC features into the MentalBERT model with label smoothing obtains better performance than MentalBERT in F1-score by 3.36\% and better performance than M-MentalBERT (LIWC) without label smoothing in F1-score by 0.63\%. However, we believe that even a small improvement can make a difference.
    \item \textit{Finding 2:} Label smoothing improves both the performance and the calibration of the proposed approaches. The calibration of the proposed approaches is measured via two metrics, namely Expected Calibration Error and Adaptive Calibration Error.
    \item \textit{Finding 3:} Findings from a linguistic analysis reveal that people in stressful and/or depressive conditions use words belonging to specific LIWC categories more frequently than others.
\end{itemize}

There are several limitations related to this study. 

\begin{itemize}
\item \textit{Hyperparameter Tuning:} Due to limited access to GPU resources, we were not able to perform hyperparameter tuning. On the contrary, we tried some combinations of parameters. We believe that the adoption of the hyperparameter tuning procedure through the access to GPU resources would increase further the classification performance. 
\item \textit{Explainability:} The present study is not accompanied with explainability techniques, i.e., Integrated Gradients \cite{10.5555/3305890.3306024}, and so on. Therefore, we aim to apply explainability techniques in the future. 
\item Due to limited access to GPU resources and similarly to prior work \cite{9904737,9733425,9447025}, we were not able to perform multiple runs for testing for statistical significance in terms of the \textit{Depression\_Mixed} and \textit{Dreaddit} datasets. 
\end{itemize}

\section{Conclusion and Future Work}
In this paper, we present a new method for identifying stress and depression in social media text by injecting linguistic information into transformer-based models. Also, it is the first study exploiting label smoothing, in order to ensure that our model is calibrated. We evaluate our proposed methods on three publicly available datasets, which include a depression detection dataset (binary classification), a stress detection dataset, and a depression detection dataset (multiclass classification - severity of depression). Findings suggest that transformer-based networks combined with linguistic information lead to performance improvement in comparison with transformer-based networks. Also, applying label smoothing yields both to the performance improvement and better calibration of the proposed models. Specifically, in terms of the Depression\_Mixed dataset, we found that the injection of top2vec features into BERT and MentalBERT models along with label smoothing obtained the highest F1-score and Accuracy. Regarding the Dreaddit dataset, results showed that the integration of LIWC features into language models based on transformers in conjunction with label smoothing yielded the highest F1-score and Accuracy. With regards to the Depression\_Severity dataset, findings showed that the injection of NRC features into the BERT model and the integration of features derived by LDA topics, namely GOSS features, into the MentalBERT model yielded the highest weighted F1-scores. We also conduct a linguistic analysis and show that stressful and depressive posts present high correlations with common LIWC categories.

In the future, we plan to exploit transfer learning and domain adaptation methods. Also, employing explainable multimodal models is one of our future plans.  In addition, we plan to exploit more methods for enhancing transformer-based models with external knowledge. Finally, we aim to contribute further to the uncertainty estimation by exploiting Monte Carlo dropout \cite{10.5555/3045390.3045502}.



\bibliographystyle{IEEEtran}  
\bibliography{IEEEabrv, references} 
 
%


\vspace{-35pt}

\begin{IEEEbiography}[{\includegraphics[width=1in,height=1.25in,clip,keepaspectratio]{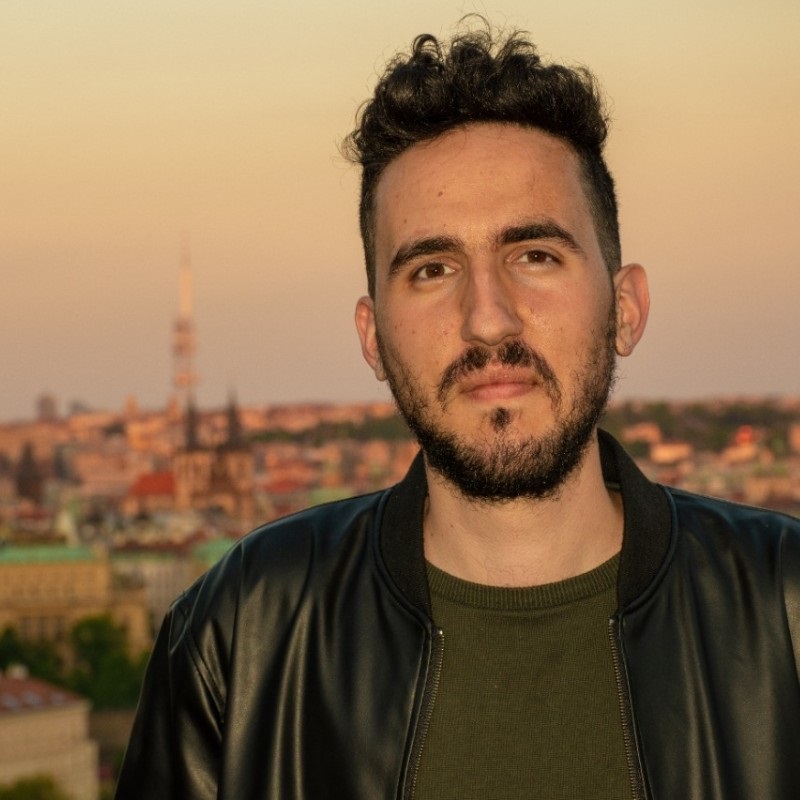}}]{Loukas Ilias} received the integrated master’s degree
from the School of Electrical and Computer Engineering (SECE), National Technical University of
Athens (NTUA), Athens, Greece, in June 2020,
where he is currently pursuing the Ph.D. degree with
the Decision Support Systems (DSS) Laboratory,
SECE. He has completed a Research Internship
with University College London (UCL), London,
U.K.
He is a Researcher with the DSS Laboratory,
NTUA, where he is involved in EU-funded research
projects. He has published in numerous journals, including IEEE Journal of Biomedical and Health Informatics, Expert Systems With Applications (Elsevier), Applied Soft Computing (Elsevier), Computer Speech and
Language (Elsevier), IEEE Access, and Frontiers in Aging Neuroscience. His
research has also been accepted for presentation at international conferences,
including the IEEE-EMBS
International Conference on Biomedical and Health Informatics (BHI’22)
and the IEEE International Conference on Acoustics, Speech, and Signal
Processing (ICASSP) 2023. His research interests include speech processing,
natural language processing, social media analysis, and the detection of
complex brain disorders.

\end{IEEEbiography}

\vspace{-36pt}

\begin{IEEEbiography}[{\includegraphics[width=1in,height=1.25in,clip,keepaspectratio]{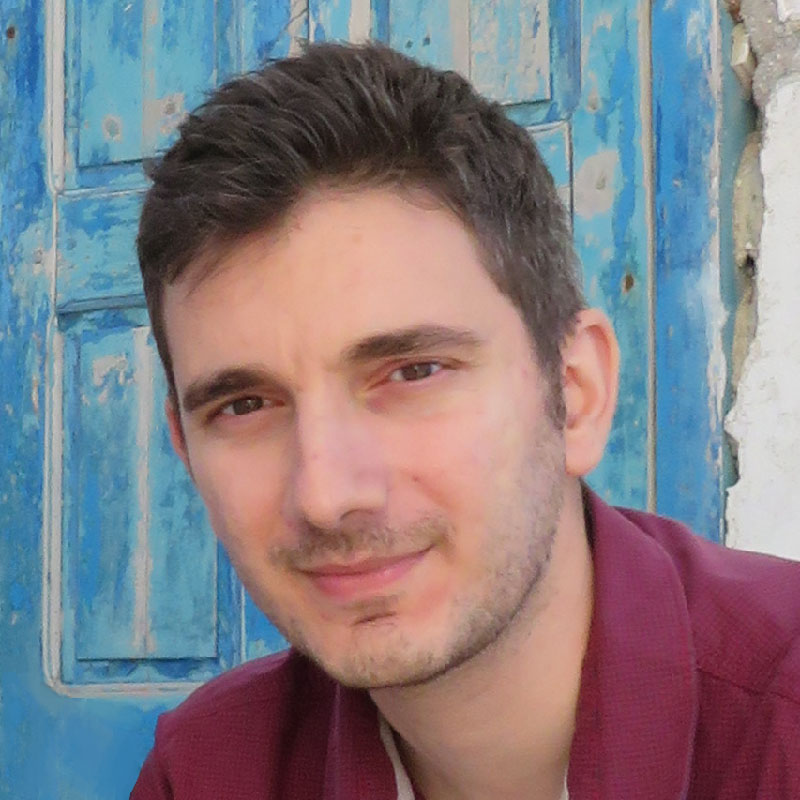}}]{Spiros Mouzakitis}
received the Ph.D. degree from
the School of Electrical and Computer Engineering,
National Technical University of Athens, Athens,
Greece, in 2009.
He has more than ten years of industry experience
in software engineering for banking, e-commerce
systems, and energy suppliers. He worked as
a Project Manager for 16 years in more than
20 projects (H2020, FP7, FP6, and FP5 research
projects) related to big data, open data, market/impact analysis in the context of ICT technologies, web development, and enterprise interoperability. He is a Senior Research
Analyst with the Decision Support Systems Laboratory, National Technical
University of Athens. He has published in numerous journals and presented
his research at international conferences. His current research is focused on
decision analysis in the field of decision support systems based on machine
learning/deep learning, big and linked data analytics, as well as optimization
systems and algorithms.

\end{IEEEbiography}

\vspace{-36pt}

\begin{IEEEbiography}[{\includegraphics[width=1in,height=1.25in,clip,keepaspectratio]{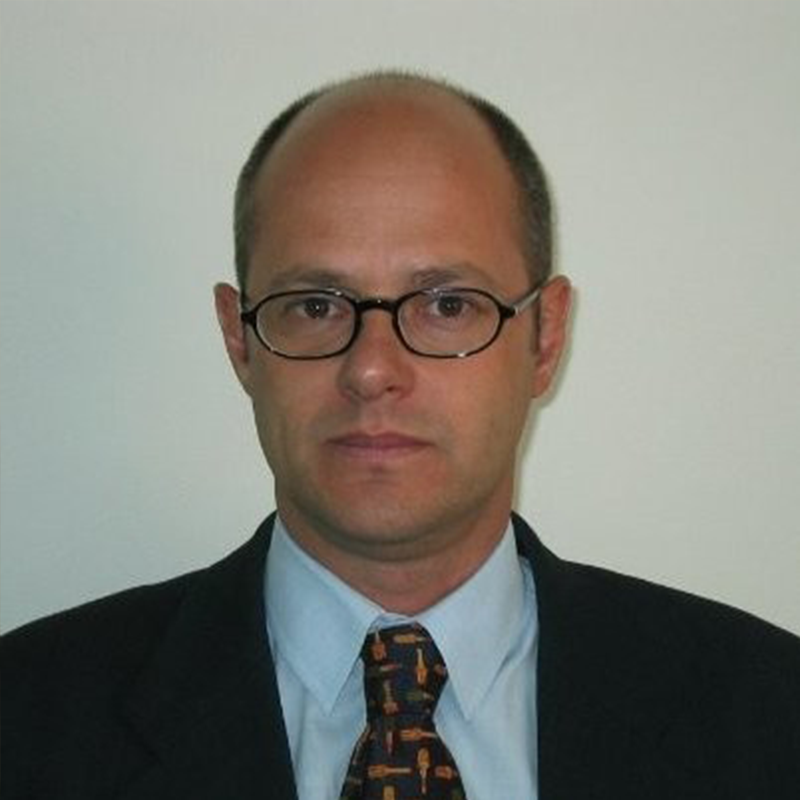}}]{Dimitris Askounis}
was the Scientific Director of
over 50 European research projects in the above
areas (FP7, Horizon2020, and so on). For a number
of years, he was an Advisor to the Minister of
Justice and the Special Secretary for Digital Convergence for the introduction of information and
communication technologies in public administration. Since June 2019, he has been the President
of the Information Society SA, Kallithea, Greece.
He is currently a Professor at the School of Electrical and Computer Engineering, National Technical
University of Athens (NTUA), Athens, Greece, and the Deputy Director
of the Decision Support Systems Laboratory. He has over 25 years of
experience in decision support systems, intelligent information systems and
manufacturing, e-business, e-government, open and linked data, big data
analytics, Artificial Intelligence (AI) algorithms, and the application of modern
Information Technology (IT) techniques in the management of companies and
organizations.
\end{IEEEbiography}


\vfill

\end{document}